\documentclass[letterpaper]{article} 
\usepackage{aaai2026}  
\usepackage{times}  
\usepackage{helvet}  
\usepackage{courier}  
\usepackage[hyphens]{url}  
\usepackage{graphicx} 
\usepackage{natbib}  
\usepackage{caption} 
\usepackage{booktabs} 
\usepackage{amsmath}
\usepackage{amssymb}

\usepackage{listings}
\usepackage{xcolor}
\usepackage{tcolorbox}
\usepackage{multirow} 
\tcbuselibrary{listingsutf8}

\usepackage{algorithm}
\usepackage{algorithmic}
\lstdefinelanguage{json}{
    basicstyle=\ttfamily\small,
    numbers=none,
    stepnumber=1,
    numbersep=8pt,
    showstringspaces=false,
    breaklines=true,
    frame=single,
    backgroundcolor=\color{gray!5},
    literate=
     *{0}{{{\color{black}0}}}{1}
      {1}{{{\color{black}1}}}{1}
      {2}{{{\color{black}2}}}{1}
      {3}{{{\color{black}3}}}{1}
      {4}{{{\color{black}4}}}{1}
      {5}{{{\color{black}5}}}{1}
      {6}{{{\color{black}6}}}{1}
      {7}{{{\color{black}7}}}{1}
      {8}{{{\color{black}8}}}{1}
      {9}{{{\color{black}9}}}{1}
      {:}{{{\color{black}:}}}{1}
      {,}{{{\color{black},}}}{1}
      {"}{{{\color{black}"}}}{1}
}
\newtcolorbox{promptbox}[1][]{
  colbacktitle=black!60,
  coltitle=white,
  fontupper=\footnotesize,
  colback=gray!5,
  boxsep=5pt,
  left=5pt,
  right=5pt,
  top=5pt,
  bottom=5pt,
  boxrule=1pt,
  title={#1},
  width=0.99\linewidth, 
}
\usepackage{newfloat}
\usepackage{listings}
\DeclareCaptionStyle{ruled}{labelfont=normalfont,labelsep=colon,strut=off} 
\floatstyle{ruled}
\newfloat{listing}{tb}{lst}{}
\floatname{listing}{Listing}
\title{Geoint-R1: Formalizing Multimodal Geometric Reasoning with Dynamic Auxiliary Constructions}

\author{
    Jingxuan Wei$^{1,3}$, Caijun Jia$^{1,3}$, Qi Chen$^{1,3}$, Honghao He$^{1,3}$, Linzhuang Sun$^{1,3}$, Conghui He$^{2}$, \\
    Lijun Wu$^{2}$, Bihui Yu$^{1,3}$, Cheng Tan$^{2}$\thanks{\ \ Corresponding author.}  \\
    $^1$Shenyang institute of computing technology, Chinese academy of sciences \\
    $^2$ Shanghai AI Laboratory 
    $^3$University of Chinese Academy of Sciences  \\
    {\tt\small weijingxuan20@mails.ucas.edu.cn, chengtan9907@gmail.com}
}

\usepackage{bibentry}

\begin{document}

\maketitle

\begin{abstract}
Mathematical geometric reasoning is essential for scientific discovery and educational development, requiring precise logic and rigorous formal verification. While recent advances in Multimodal Large Language Models (MLLMs) have improved reasoning tasks, existing models typically struggle with formal geometric reasoning, particularly when dynamically constructing and verifying auxiliary geometric elements. To address these challenges, we introduce Geoint-R1, a multimodal reasoning framework designed to generate formally verifiable geometric solutions from textual descriptions and visual diagrams. Geoint-R1 uniquely integrates auxiliary elements construction, formal reasoning represented via Lean4, and interactive visualization. To systematically evaluate and advance formal geometric reasoning, we propose the Geoint benchmark, comprising 1,885 rigorously annotated geometry problems across diverse topics such as plane, spatial, and solid geometry. Each problem includes structured textual annotations, precise Lean4 code for auxiliary constructions, and detailed solution steps verified by experts. Extensive experiments demonstrate that Geoint-R1 significantly surpasses existing multimodal and math-specific reasoning models, particularly on challenging problems requiring explicit auxiliary element constructions. 
\end{abstract}


\section{Introduction}

Mathematical geometric reasoning is fundamental for cognitive, educational, and scientific development, requiring precise logical inference and formal verification. Recent advancements in MLLMs have shown promising results on various reasoning tasks~\cite{yang2023mm,tan2024boosting,wei2024enhancing,wu2023multimodal}. However, existing approaches primarily focus on raw textual or visual geometry problems, lacking capabilities in rigorous reasoning and failing to dynamically construct and visualize necessary auxiliary elements in complex multimodal geometric scenarios~\cite{lumathvista,qiao2024we,zhang2024mathverse}.

\begin{figure*}[t]
\centering
\includegraphics[width=\linewidth]{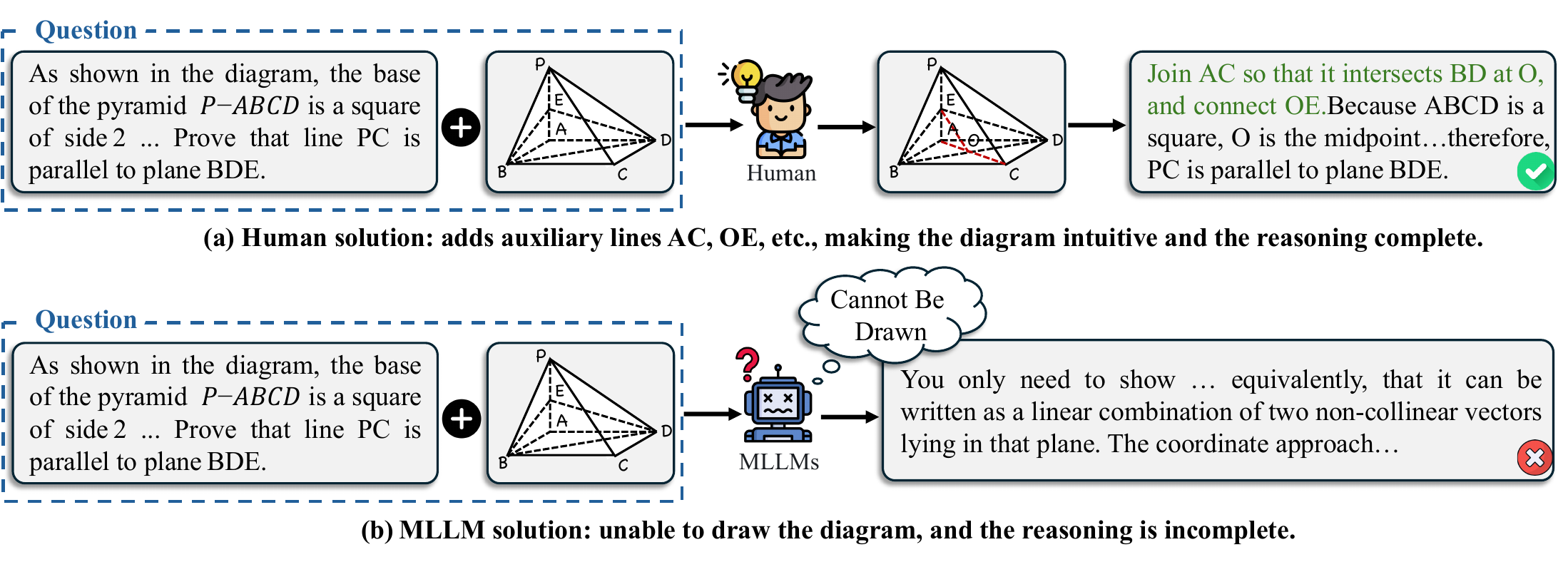}
\caption{Comparison between human and MLLM solutions to geometric problems requiring auxiliary constructions.}
\label{fig:introduction}
\end{figure*}

We introduce the formal geometric reasoning task, which involves generating complete and verifiable geometric solutions from multimodal inputs consisting of textual descriptions and geometric diagrams. As illustrated in Figure~\ref{fig:introduction}, this task fundamentally differs from existing geometric tasks by requiring not only semantic correctness but also the identifying, formal constructing, and visually demonstrating auxiliary elements. This introduces specific core challenges: \textbf{(1) accurately constructing auxiliary lines essential to solving geometric problems, and (2) generating formally verifiable reasoning steps supported by visual feedback that clearly aligns with geometric structures and user intent}.

To facilitate progress in formal multimodal geometric reasoning, we introduce \textbf{Geoint}, a comprehensive benchmark dataset explicitly designed for formal geometric problem-solving. Geoint encompasses 1,885 carefully curated geometric questions across diverse categories including plane, spatial, and solid geometry problems. Each problem is richly annotated with both structured textual descriptions and accompanying visual diagrams to support multimodal understanding. Furthermore, Geoint leverages the Lean 4 proof assistant to formally represent geometric elements and relationships, enabling rigorous and complete formal reasoning within a verifiable framework.

Building upon the Geoint benchmark, we propose \textbf{Geoint-R1}, a geometric reasoning framework characterized by its capability to generate formally verifiable geometric solutions with dynamic auxiliary constructions. Geoint-R1 employs a two-stage training paradigm combining supervised fine-tuning with reinforcement learning guided by a verification reward model that assigns rewards based on correctness, accuracy of auxiliary constructions, and adherence to structured output formats. Additionally, a curriculum learning strategy further enhances the model's ability to handle increasingly complex geometric reasoning tasks.

Our main contributions are summarized as follows:
\begin{itemize}
\item We introduce the formal geometric reasoning task along with the Geoint benchmark, a rigorously annotated dataset uniquely integrating structured textual annotations and visual auxiliary line constructions within a complete and verifiable framework.
\item We propose the Geoint-R1 framework, an advanced multimodal geometric reasoning model featuring automated identification, construction, and visualization of auxiliary lines within a unified SFT and RL-based pipeline.
\item Extensive experiments demonstrate the superior performance of Geoint-R1 compared to state-of-the-art multimodal reasoning models, validating the efficacy of formal auxiliary construction, verification reward model, and training strategies.
\end{itemize}

\section{Related Work}
\subsection{Mathematical Benchmarks}
Recent advances in mathematical reasoning benchmarks have produced a series of multimodal datasets for evaluating large models on mathematical problems with visual content. MathVista~\cite{lumathvista}, We-Math~\cite{qiao2024we}, MathVerse~\cite{zhang2024mathverse}, and Polymath~\cite{gupta2024polymath} cover a broad spectrum of topics ranging from plane and solid geometry to pattern recognition, and adopt diverse formats such as multiple-choice, open-ended, and proof-style questions. Many of these benchmarks, emphasize stepwise evaluation and fine-grained error analysis, while MathScape~\cite{zhou2024mathscape}, MM-MATH~\cite{sun2024mm}, and MMSciBench~\cite{ye2025mmscibench} further introduce hierarchical or process-based assessments, supporting deeper insight into multi-step reasoning and concept coverage. For spatial and geometric reasoning, SolidGeo~\cite{wang2025solidgeo} provides a collection of annotated three-dimensional geometry problems that enable analysis of spatial understanding. However, these datasets do not provide formali representations that support stepwise visual reasoning or interactive geometric constructions within a complete framework. In contrast, our work constructs a benchmark that links geometry problem to formalized reasoning steps and visualizations of auxiliary constructions, enabling explicit modeling of the reasoning process through the dynamic auxiliary constructions. 

\subsection{Visual Reasoning}
Visual reasoning has become a core capability for multimodal models, enabling them to perform complex tasks that require iterative image manipulation. Recent works such as MM-REACT~\cite{yang2023mm}, VisuoThink~\cite{wang2025visuothink}, CMMCoT~\cite{zhang2025cmmcot}, VGR~\cite{wang2025vgr}, Seg-Zero~\cite{liu2025seg}, and VisionReasoner~\cite{liu2025visionreasoner} propose diverse frameworks for integrating vision-language models, tool-use, and reinforcement learning to improve visual question answering, multi-image reasoning, and structural scene understanding. Methods like MMFactory~\cite{fan2024mmfactory}, DiagramAgent~\cite{wei2025words}, MathCoder-VL~\cite{wang2025mathcoder}, and PyVision~\cite{zhao2025pyvision} introduce agent-based or programmatic approaches to visual manipulation, supporting flexible tool selection, diagram editing, and code-based image operations. Other approaches leverage explicit reasoning chains and learning paradigms, as seen in PROGRM~\cite{zhang2025progrm}, GoT~\cite{fang2025got}, Show-o2~\cite{xie2025show}, ControlThinker~\cite{han2025controlthinker}, and T2I-R1~\cite{jiang2025t2i}, which further enhance visual generation, multi-step planning, and semantic control through supervised or reinforcement learning strategies. However, they are not designed for precise mathematical problems and do not support stepwise auxiliary construction or visual feedback in formal geometric reasoning.
\section{Methodology}

\begin{figure*}[t]
    \centering
    \includegraphics[width=\linewidth]{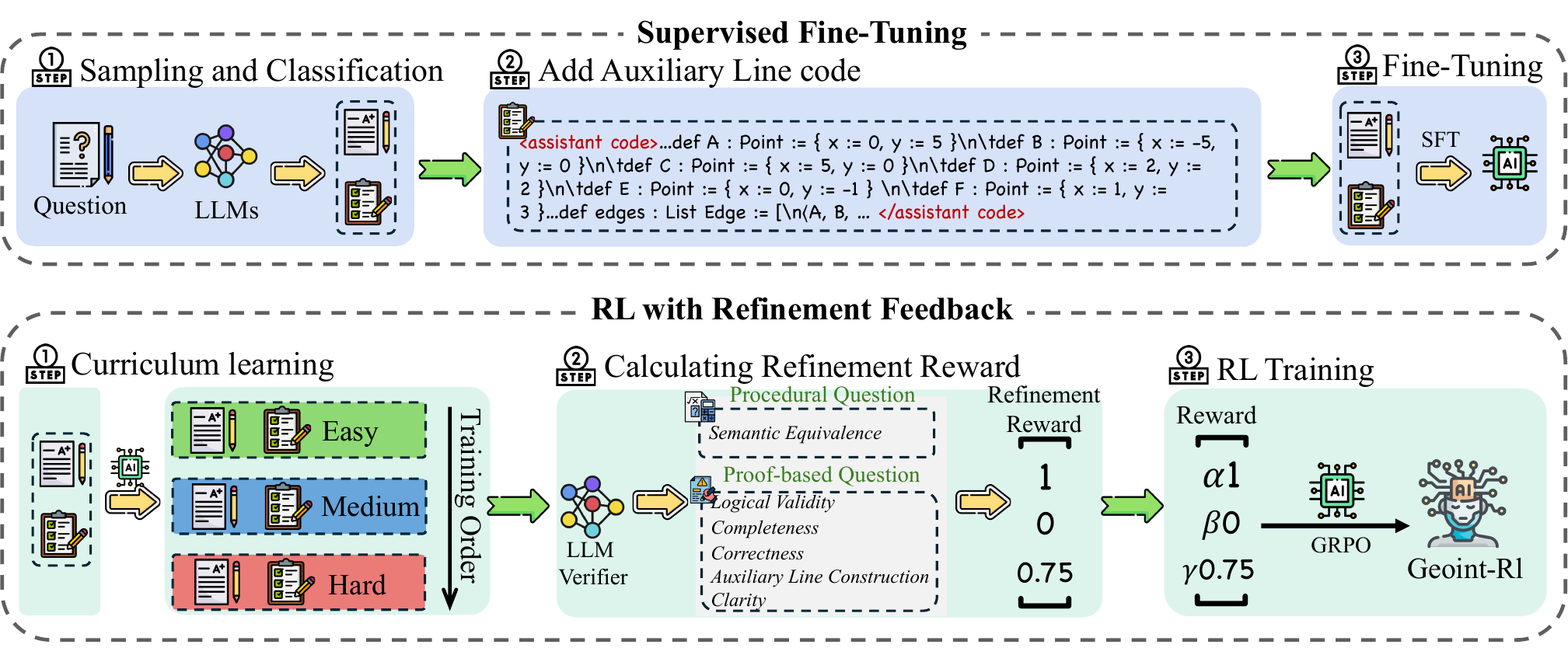}
    \caption{The Geoint-R1 architecture and training pipeline.}
    \label{fig:geoint_pipeline}
\end{figure*}

Our goal is to develop a model $\mathcal{F}_\theta$ that, given a textual description $T_i$ and a visual diagram $I_i$of a geometry problem, generates a complete and formally verifiable solution. A solution consists of a sequence of reasoning steps $P_i$, a final answer $A_i$, and, when necessary, a set of auxiliary constructions $C_i$ represented in a formal language. Formally, the model predicts the tuple $(P_i, C_i, A_i)$:
\begin{equation}
    (P_i, C_i, A_i) = \mathcal{F}_\theta(T_i, I_i).
\end{equation}
To achieve this, we introduce Geoint-R1, a framework trained in two main stages: Supervised Fine-Tuning (SFT) and Reinforcement Learning with an verification reward model. The overall architecture is illustrated in Figure \ref{fig:geoint_pipeline}.

\subsection{Stage 1: Supervised Fine-Tuning}

The initial stage aims to teach the base model the fundamental structure of geometric solutions and the syntax of formal constructions. Given the training dataset $\mathcal{D}=\{(T_i, I_i, P_i^{ref}, C_i^{ref},  A_i^{ref})\}$, where superscripts denote the reference solution, the objective is to maximize the likelihood of generating the correct solution.

The model $\mathcal{F}_\theta$ is fine-tuned by minimizing the standard negative log-likelihood (NLL) loss:
\begin{equation}
\mathcal{L}_{\text{SFT}}(\theta) = - \sum_{i=1}^{N} \log P(P_i^{ref}, C_i^{ref}, A_i^{ref} | T_i, I_i; \theta),
\end{equation}
where for problems without auxiliary lines we simply set $C_i^{ref}=\emptyset$. This step uses standard teacher-forcing to ground the model in correct proof structure and Lean4 syntax.

\subsection{Stage 2: RL with Verification Reward Model}

While SFT provides a strong foundation, it may not sufficiently optimize for semantic correctness or logical rigor. The second stage refines the policy of the SFT model, denoted as $\pi_\theta$, using reinforcement learning guided by a predefined verification reward model. First, we re-apply the same reject-sampling procedure on the SFT outputs to curate a training set where the model still has room for improvement. Each problem in this curated set receives a difficulty score $d_i=1-\frac{1}{K}\sum_k \delta_{i,k}$ where $\delta_{i,k}$ is the indicator function for the $k$-th candidate solution being correct. We sort problems by ascending $d_i$, implementing a curriculum learning in each sampling that presents easier instances before harder ones. 

The RL objective is to maximize the expected reward:
\begin{equation}
\mathcal{J}(\theta) = \mathbb{E}_{(T_i, I_i) \sim \mathcal{D}} \left[ \mathbb{E}_{(P_i, C_i, A_i) \sim \pi_{\theta}(\cdot|T_i, I_i)} [R(P_i, C_i, A_i)] \right].
\end{equation}
At each step, a single roll-out is sampled for a problem, scored by our verification model into a scalar reward $R(\cdot)$, and used to update $\theta$ via GRPO~\cite{shao2024deepseekmath}. The curriculum ordering ensures stable progression from simpler proof patterns to more complex auxiliary constructions.

\subsection{Verification Reward Model}

We design a reward function $R(\cdot)$ that evaluates model outputs with fine-grained feedback on solution quality. Concretely, $R(\cdot)$ is a weighted sum of three components:
\paragraph{Correctness} For answer-based questions, the accuracy $\mathrm{acc}\in\{0,1\}$ is computed by exactly matching with the reference answer. For proof-based questions, the accuracy $\mathrm{acc}\in[0,1]$ is computed by the GPT-4o's evaluation. We further modulate the correctness score by the accuracy of the auxiliary line construction:
\begin{equation}
F_{corr}(x) = \begin{cases}
\min(1,\mathrm{acc} + \lambda),&\text{if auxiliary are correct} \\
\max(0, \mathrm{acc} - \lambda),&\text{otherwise}
\end{cases}
\end{equation}
where the constant $\lambda$ boosts or penalizes the continuous score depending on auxiliary-line correctness.

\paragraph{Auxiliary Line} We award $F_{aux}=1$ whenever the model exactly matches the reference set of auxiliary constructions (i.e.\ all needed lines are both identified and precisely encoded in Lean4), and $F_{aux}=0$ otherwise.

\paragraph{Format} To ensure the model's responses can be unambiguously parsed by downstream systems, we require a strict think-answer wrapper. $F_{fmt}(x) = 1$ if the output format matches the required template; $0$ otherwise.

We then combine these sub-rewards into a single scalar:
\begin{equation}
    R(x) = \alpha F_{corr}(x) + \beta F_{aux}(x) + \gamma F_{fmt}(x),
\end{equation}
where $\alpha$, $\beta$, and $\gamma$ are hyperparameters balancing the different aspects of solution quality.

\section{Geoint Dataset}

\begin{figure*}[t]
    \centering
    \includegraphics[width=0.95\linewidth]{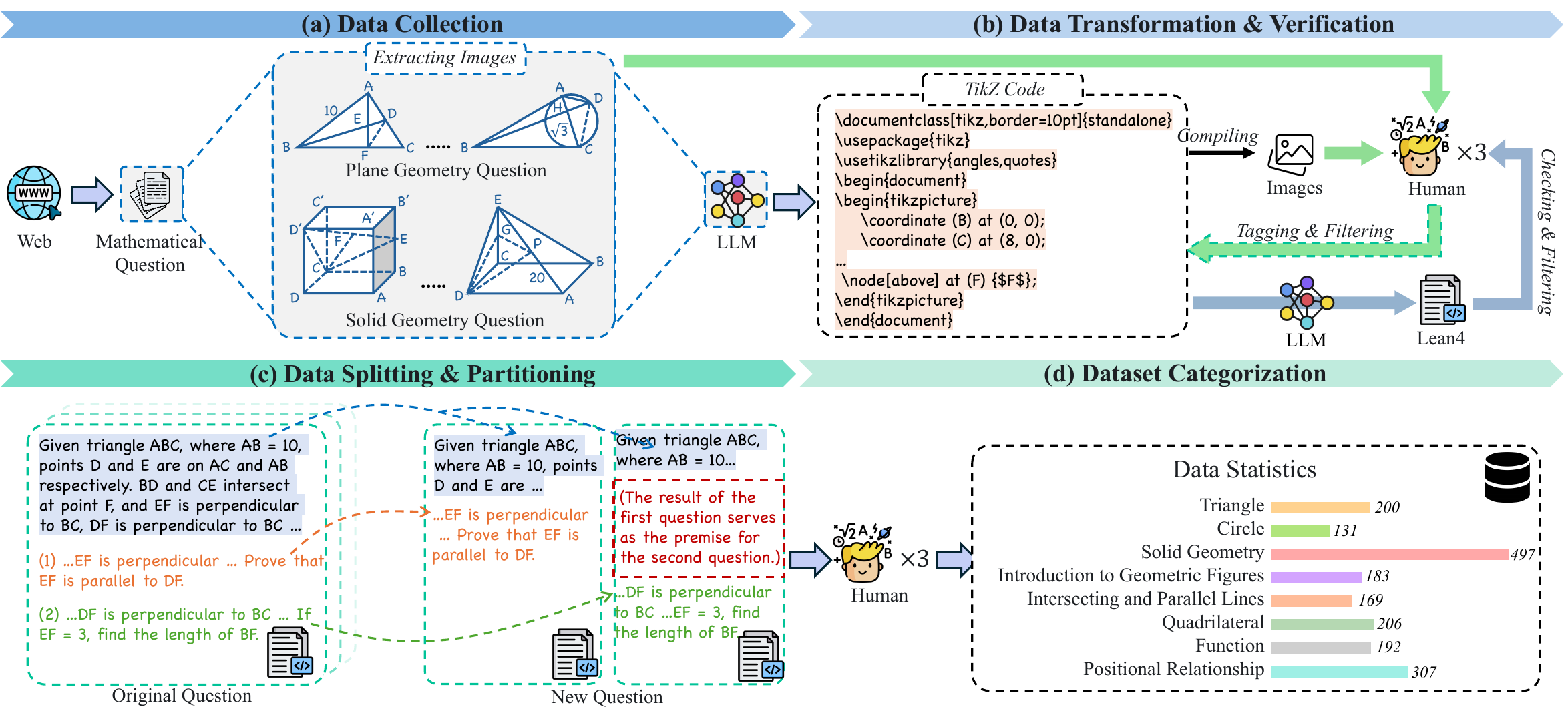}
    \caption{The overview of the Geoint dataset construction pipeline.}
    \label{fig:data_pipeline}
\end{figure*}

We present \textbf{Geoint}, a multimodal benchmark designed for formal and interactive geometric problem solving. Geoint comprises a diverse set of geometry questions. A distinguishing feature of Geoint is its detailed annotation of auxiliary constructions—such as dynamically added auxiliary lines. At each geometric reasoning step, enabling explicit modeling and visualization of the problem-solving process. The dataset construction pipeline is illustrated in Figure~\ref{fig:data_pipeline}. 

\paragraph{Data Collection}
We begin by sourcing a wide range of geometry problems from publicly available web resources. All problems are manually reviewed and categorized by knowledge type into two main domains: plane and solid geometry.

\paragraph{Data Transformation \& Verification}
For each problem, we extract associated images and convert diagrams into TikZ code. The TikZ code is compiled into diagram and compared against the original, with quality rated by human annotators. For high-quality cases, we employ the DeepSeek-R1~\cite{guo2025deepseek} to translate TikZ code into Lean4 formal proofs, which are then subject to manual verification for correctness. The resulting Lean4 code and corresponding TikZ diagrams are attached to each problem's textual description to support formalized reasoning and visualization.

\paragraph{Data Splitting \& Partitioning}
Each original geometry problem is decomposed into individual question-answer pairs, splitting multi-step problems into separate entries following a ``prompt + question'' structure. For multi-part problems where one sub-question depends on the result of the previous question, we explicitly include prior conclusions as context in the prompt for the subsequent sub-question.

\paragraph{Dataset Categorization}
The final dataset is categorized into nine knowledge types: Triangle, Solid Geometry, Quadrilateral, Function, Positional Relationship, Circle, Geometry Figures, and Intersecting/Parallel Lines. Each entry is double-checked for category accuracy and quality.

\paragraph{Expert Annotation}
Throughout the pipeline, all manual validation and annotation are conducted by three mathematics professionals with master degree, ensuring the integrity and mathematical soundness of the dataset.

\subsection{Data Analysis}

Geoint contains 1,885 question-answer pairs. Table~\ref{tab:sample_stats} summarizes the overall sample counts, training/testing splits, and question type distribution for each subset. Table~\ref{tab:knowledge_length_stats} reports the distribution across knowledge categories, as well as token statistics for question, answer, and code. These analyses highlight the diversity of Geoint, including the presence of both answer-based and proof-based questions, and a wide range of diagrammatic and formalized solution lengths.

\begin{table}[t]
  \footnotesize
  \centering
  \caption{Sample statistics for Geoint.}
    \setlength{\tabcolsep}{2mm}{
    \begin{tabular}{lcc}
    \toprule
    \textbf{Statistic} & \textbf{Auxiliary line} & \textbf{No auxiliary line} \\
    \midrule
    Total samples & 429 & 1456 \\
    Training samples & 325 & 1289 \\
    Testing samples & 104 & 167 \\
    \midrule
    \multicolumn{3}{l}{\textbf{Question type}} \\
    Answer-based questions & 241 & 956 \\
    Proof-based questions & 188 & 500 \\
    \bottomrule
    \end{tabular}
    }
  \label{tab:sample_stats}
\end{table}

\begin{table}[t]
\footnotesize
\centering
\caption{Knowledge type and token statistics for Geoint.}
    \setlength{\tabcolsep}{1.5mm}{
    \begin{tabular}{lcc}
    \toprule
    \textbf{Knowledge type} & \textbf{Auxiliary line} & \textbf{No auxiliary line} \\
    \midrule
    Triangle & 50 & 150 \\
    Solid Geometry & 118 & 379 \\
    Quadrilateral & 54 & 152 \\
    Function & 20 & 172 \\
    Positional Relationship & 90 & 217 \\
    Circle & 38 & 93 \\
    Geometric Figures & 26 & 157 \\
    Intersecting/Parallel Lines & 33 & 136 \\
    \midrule
    \multicolumn{3}{l}{\textbf{Question length (tokens)}} \\
    Minimum & 30 & 29 \\
    Maximum & 1559 & 4104 \\
    Average & 173.92 & 156.79 \\
    \midrule
    \multicolumn{3}{l}{\textbf{Answer length (tokens)}} \\
    Minimum & 88 & 5 \\
    Maximum & 2558 & 4589 \\
    Average & 468.38 & 267.2 \\
    \midrule
    \multicolumn{3}{l}{\textbf{Code length (tokens)}} \\
    Minimum & 41 & 0 \\
    Maximum & 1149 & 0 \\
    Average & 426.31 & 0 \\
    \bottomrule
    \end{tabular}
    }
\label{tab:knowledge_length_stats}
\end{table}

Geoint exhibits a balanced representation of both auxiliary/non-auxiliary line problems across diverse geometric topics. The average answer and code length for auxiliary line problems are significantly higher, reflecting the complexity of formalized visual reasoning. This fine-grained structure provides a strong foundation for evaluating both visual and formal reasoning in multimodal models.

\subsection{Evaluation Metrics}

We employ tailored automatic evaluation metrics to rigorously assess both answer- and proof-based geometric reasoning within Geoint. For each sample, all evaluations are conducted using the DeepSeek-V3~\cite{liu2024deepseek}, ensuring consistent and reproducible results.

For \textbf{answer-based questions}, we extract the final answer from the model's output and compare it with the reference solution to compute accuracy. The evaluation follows a strict grading prompt, considering mathematical equivalence and only awarding credit for correct final results. The scoring mechanism outputs 1 point for correct or equivalent answers, and 0 otherwise. For \textbf{proof-based questions}, we adopt a comprehensive process-based evaluation inspired by DeepTheorem~\cite{zheng2024deeptheorem}. Each proof is automatically scored along five weighted criteria: logical validity (30\%), completeness (20\%), correctness (20\%), auxiliary line construction (20\%), and clarity (10\%). Sub-scores for each dimension are aggregated into a total score in $[0,1]$. The prompts used for automated evaluation are shown in Figure~\ref{fig:qa_prompt} and Figure~\ref{fig:proof_prompt} in the Appendix.

\section{Experiment}

\paragraph{Setup.} In the SFT, we perform full-parameter updates while freezing the vision encoder and multi-modal projector to preserve visual capacity. The maximum input length is set to 4096 tokens, with a per-device batch size of 4, gradient accumulation of 2, and a learning rate of 1e-5 scheduled by cosine annealing with 10\% warmup. Training runs for 4 epochs with bf16 mixed precision and DeepSpeed ZeRO-3 parallelization; checkpoints are saved every 100 steps and loss is logged every 10 steps. Reinforcement learning is applied to further optimize the SFT model using rule-based policy optimization. Training data is curated by rejection sampling, generating 8 candidate outputs per sample and filtering trivial or unsolvable cases. Both training stages are implemented using Qwen2.5-VL-7B~\cite{bai2025qwen2}, with RL accelerated by FlashAttention v2 and distributed across 8 GPUs via torchrun and DeepSpeed ZeRO-3. 

\paragraph{Model}
For comprehensive benchmarking, we compare our approach against a suite of state-of-the-art multimodal and math-specific models, including both open-source and closed-source ones. The open-source multimodal models include Phi-3.5-V-4B~\cite{abdin2024phi}, LLaVA-v1.6-7B~\cite{liu2024improved}, InternLM-XComposer-2.5-7B~\cite{zhang2024internlm}, Yi-VL-6B~\cite{young2024yi}, InternVL3-8B~\cite{zhu2025internvl3}, and Qwen-VL-2.5-7B~\cite{bai2025qwen2}. Closed-source models consist of GPT-4o~\cite{hurst2024gpt}, Gemini-1.5-pro~\cite{team2024gemini}, and Gemini-1.5-flash~\cite{team2024gemini}. In addition, we include specialized math reasoning models such as MMR1-Math-v0-7B~\cite{MMR1-Math2025} and Math-LLaVA-13B~\cite{shi2024math}. These models are selected based on their recent advances and strong performance in related multimodal and mathematical reasoning tasks, serving as key baselines.

\subsection{Main Results}

\begin{table}[t]
  \footnotesize
  \centering
  \caption{Comparison of overall accuracy (\%) on answer- and proof-based geometric reasoning tasks.}
    \setlength{\tabcolsep}{2.3mm}{
    \begin{tabular}{lccc}
    \toprule
    Model & Answer   & Proof   & Average \\
    \midrule
    Phi-3.5-V-4B & 2.76  & 18.89 & 10.83 \\
    LLaVA-v1.6-7B & 8.71  & 36.80 & 22.76 \\
    InternLM-XComposer-2.5-7B & 14.02 & 41.45 & 27.74 \\
    Yi-VL-6B & 7.93  & 29.81 & 18.87 \\
    InternVL3-8B & 42.68 & 71.92 & 57.30 \\
    Qwen-VL-2.5-7B & 39.94 & 69.57 & 54.76 \\
    \midrule
    GPT-4o & 42.68 & \textbf{77.99} & 60.34 \\
    Gemini-1.5-pro & 51.53 & 73.50 & 62.52 \\
    Gemini-1.5-flash & 43.59 & 66.76 & 55.18 \\
    \midrule
    MMR1-Math-v0-7B & 49.08 & 71.93 & 60.51 \\
    Math-LLaVA-13B & 10.87 & 33.83 & 22.35 \\
    \midrule
    Geoint-R1 & \textbf{57.01} & 72.43 & \textbf{64.72} \\
    \bottomrule
    \end{tabular}}
  \label{tab:main_results}
\end{table}

Table~\ref{tab:main_results} presents the accuracy of all evaluated models on answer- and proof-based geometric reasoning tasks. At a macro level, Geoint-R1 achieves the highest overall accuracy (64.72\%), while most open-source vision-language models remain below 30\%, reflecting the general difficulty of the benchmark. For answer-based questions, Geoint-R1 leads with 57.01\%, and Gemini-1.5-pro, GPT-4o, and MMR1-Math-v0-7B also perform well, all exceeding 49\%. In contrast, proof-based questions are more challenging: GPT-4o achieves the highest accuracy (77.99\%), with Geoint-R1 and Gemini-1.5-pro both above 72\%, and most open-source models below 42\%. Although Geoint-R1 does not surpass the largest proprietary models on proof-based tasks, it remains highly competitive despite its much smaller parameter size (7B), suggesting that model scale may be a factor in achieving top performance on complex proofs. Overall, Geoint-R1 delivers the best average accuracy across both question types, indicating its balanced effectiveness on this challenging multimodal benchmark.

\subsection{Results on Auxiliary Line Problems}

\begin{figure}[h]
    \centering
    \vspace{-4mm}
    \includegraphics[width=\linewidth]{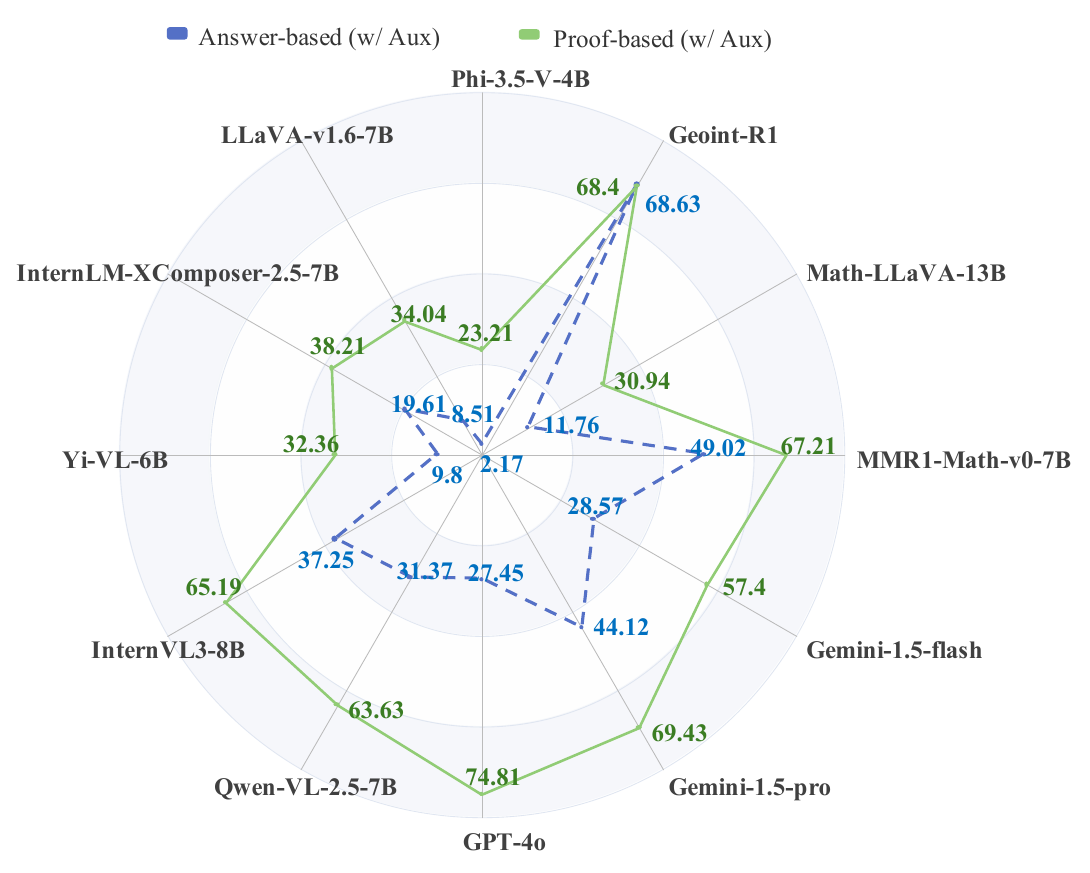} 
    \vspace{-4mm}
    \caption{Accuracy (\%) on auxiliary line problems.}
    \vspace{-4mm}
    \label{fig:auxline_results}
\end{figure}

As shown in Figure~\ref{fig:auxline_results}, most models struggle on problems that require auxiliary line construction, with open-source models generally achieving less than 40\% accuracy. For answer-based questions, Geoint-R1 achieves the highest score (68.63\%), outperforming all other models by a notable margin. In proof-based questions, Geoint-R1 achieves 68.40\%, remaining highly competitive and comparable to Gemini-1.5-pro (69.43\%), MMR1-Math-v0-7B (67.21\%), and just below GPT-4o (74.81\%), despite using far fewer parameters. These results underscore Geoint-R1's robust performance in complex geometric reasoning tasks that require explicit visual reasoning capability for both formal answer- and proof-based problems.

\subsection{Results on Non-Auxiliary Line Problems}

\begin{figure}[h]
    \centering
    \vspace{-4mm}
    \includegraphics[width=\linewidth]{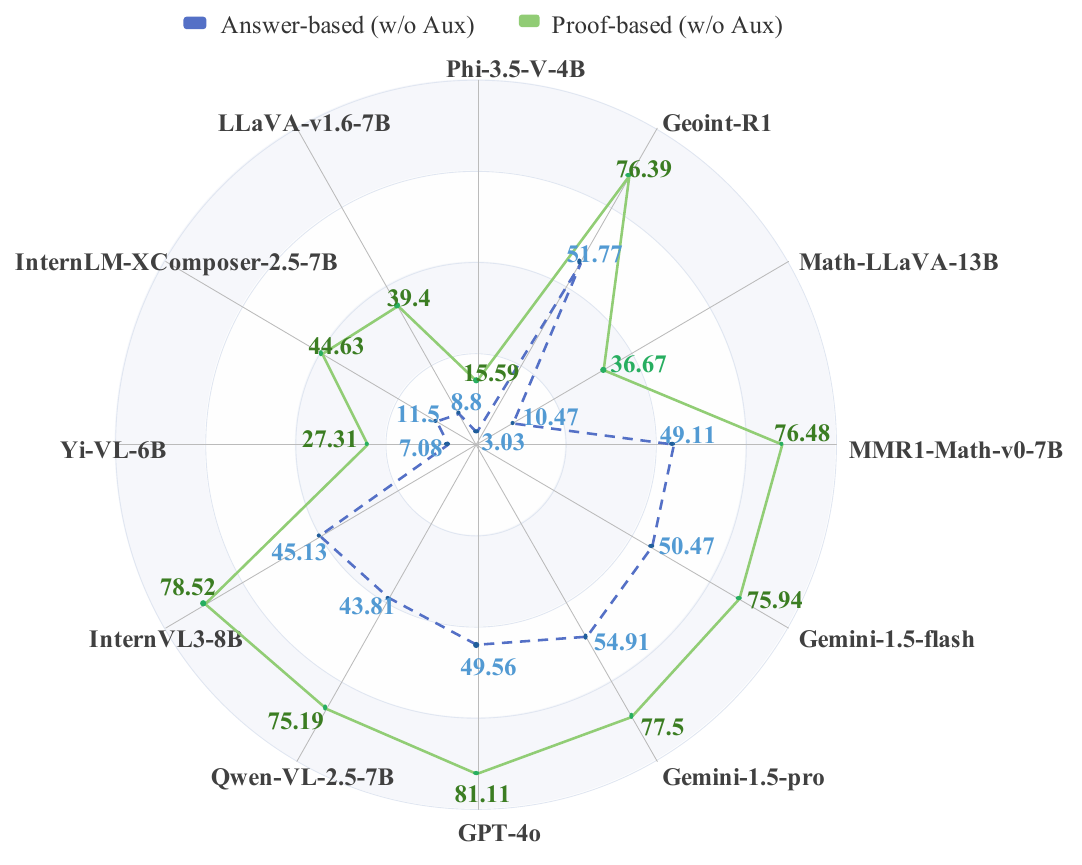} 
    \vspace{-4mm}
    \caption{Accuracy (\%) on non-auxiliary line problems.}
    \vspace{-4mm}
    \label{fig:noauxline_results}
\end{figure}

As shown in Figure~\ref{fig:noauxline_results}, all models achieve higher accuracy on non-auxiliary line problems than on auxiliary line tasks, likely because these questions are less complex and do not require explicit geometric constructions. Most closed-source and leading open-source models exceed 45\% on answer-based questions, with Geoint-R1 reaching 51.77\%, closely matching Gemini-1.5-pro (54.91\%), GPT-4o (49.56\%), and MMR1-Math-v0-7B (49.11\%). For proof-based questions, Geoint-R1 achieves 76.39\%, among the top performers together with Gemini-1.5-pro, GPT-4o, and several other math-focused models. While open-source baselines still lag on answer-based questions, the gap is reduced here.

\subsection{Ablation Study}

Table~\ref{tab:ablation} presents the ablation results for Geoint-R1 by removing LLM Verify, reinforcement learning (RL), and curriculum learning (CL) modules. At the macro level, Geoint-R1 achieves the highest average accuracy, and removing any single module leads to a significant drop. The effect is most pronounced when verification reward is removed, especially on auxiliary line problems. On proof-based tasks, Geoint-R1 achieves an average of 72.43\%, with all ablated models falling behind, again with the largest drop observed after removing verification reward. Both auxiliary and non-auxiliary line cases show consistent performance degradation across all ablations, demonstrating that these components are each essential. These findings confirm the effectiveness of our modular design for complex geometric reasoning, with each component contributing meaningfully to both answer- and proof-based performance.

\begin{table}[h]
    \centering
    \footnotesize
    \vspace{-2mm}
    \caption{Ablation results (\%) of Geoint-R1.}
    \vspace{-2mm}
      \setlength{\tabcolsep}{0.2mm}{
      \begin{tabular}{l|ccc|ccc|c}
      \toprule
      \multirow{2}{*}{Model} & \multicolumn{3}{c|}{Answer-based} & \multicolumn{3}{c|}{Proof-based} & \multirow{2}{*}{Average} \\
           & w/ Aux & w/o Aux & Avg & w/ Aux & w/o Aux & Avg &  \\
      \midrule
      Geoint-R1 & \textbf{68.63} & \textbf{51.77} & \textbf{57.01} & \textbf{68.40} & \textbf{76.39} & \textbf{72.43} & \textbf{64.72} \\
      -w/o Verify & 32.35 & 47.35 & 42.68 & 64.15 & 73.52 & 68.88 & 55.78 \\
      -w/o RL & 37.25 & 46.46 & 43.60 & 62.55 & 71.48 & 67.06 & 55.33 \\
      -w/o CL & 38.24 & 49.12 & 45.73 & 63.30 & 72.41 & 67.90 & 56.82 \\
      \bottomrule
      \end{tabular}
      }
    \label{tab:ablation}
    \vspace{-4mm}
\end{table}   

\subsection{Case Studies}

\begin{figure}[h]
    \centering
    \includegraphics[width=\linewidth]{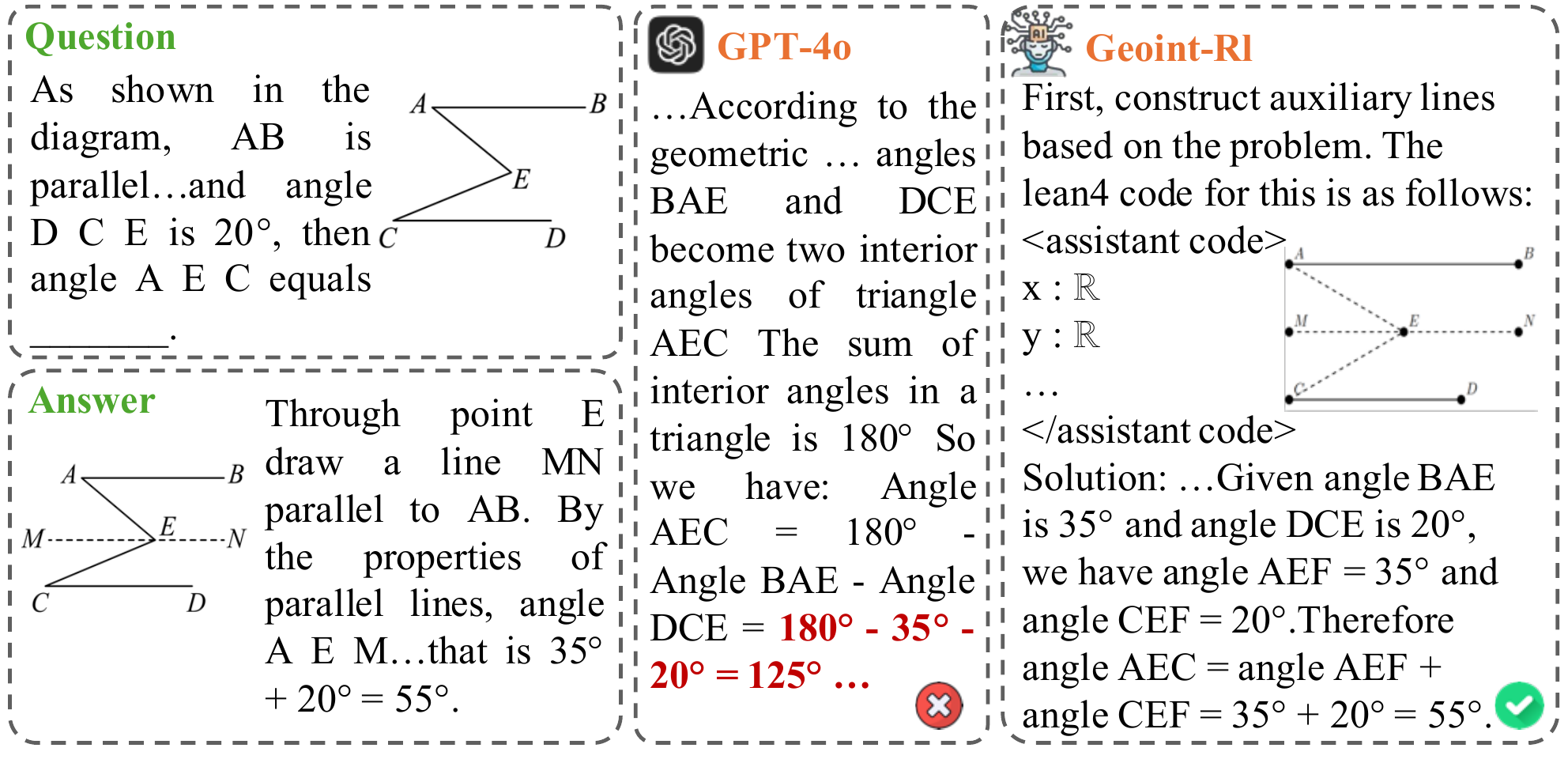}
    \vspace{-4mm}
    \caption{Auxiliary lines support accurate planar inference.}
    \vspace{-2mm}
    \label{fig:case2}
\end{figure}

We present two representative examples to demonstrate Geoint-R1’s advantage in auxiliary line construction and formal reasoning as follows:

\noindent\textbf{Example 1 (Answer-based):} As shown in Figure~\ref{fig:case2}, Geoint-R1 constructs a parallel through $E$ and directly applies angle properties to compute $\angle AEC$, while GPT-4o ignores auxiliary lines and arrives at an incorrect result.
These results highlight the necessity of explicit auxiliary construction for robust and interpretable geometric reasoning.

\begin{figure}[h]
    \centering
    \vspace{-1mm}
    \includegraphics[width=\linewidth]{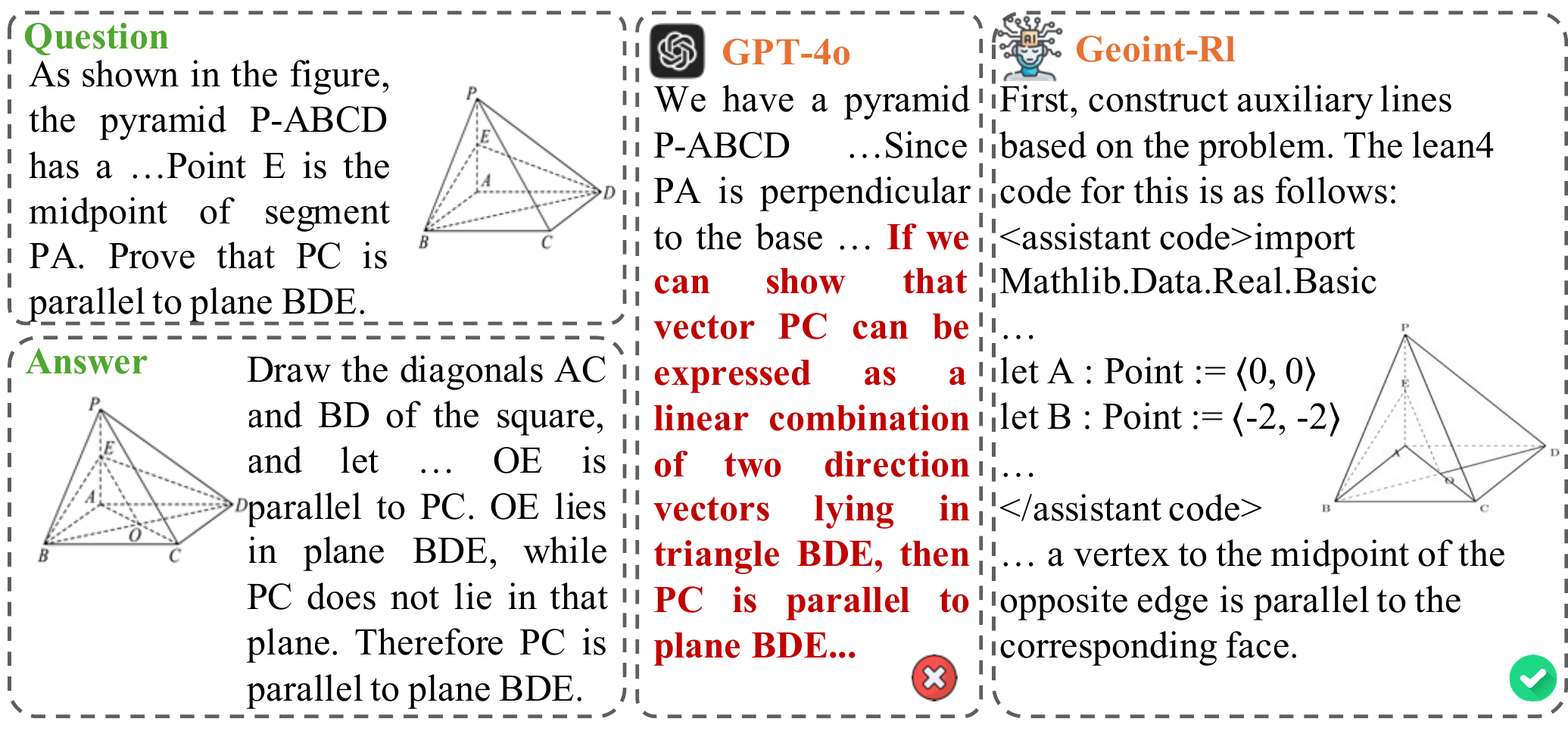}
    \vspace{-5mm}
    \caption{Auxiliary construction for spatial proof.}
    \vspace{-4mm}
    \label{fig:case1}
\end{figure}

\noindent\textbf{Example 2 (Proof-based):} Figure~\ref{fig:case1} considers a spatial proof requiring demonstration that $PC \parallel$ plane $BDE$ in a pyramid. Geoint-R1 solves this by constructing the intersection point $O$ of diagonals, connecting $O$ to $E$, and proving $OE \parallel PC$, thus delivering a concise and verifiable proof. In contrast, GPT-4o employs vector operations but fails to establish the required geometric relationships.

\subsection{Error Analysis}

\begin{figure}[h]
    \centering
    \vspace{-2mm}
    \includegraphics[width=\linewidth]{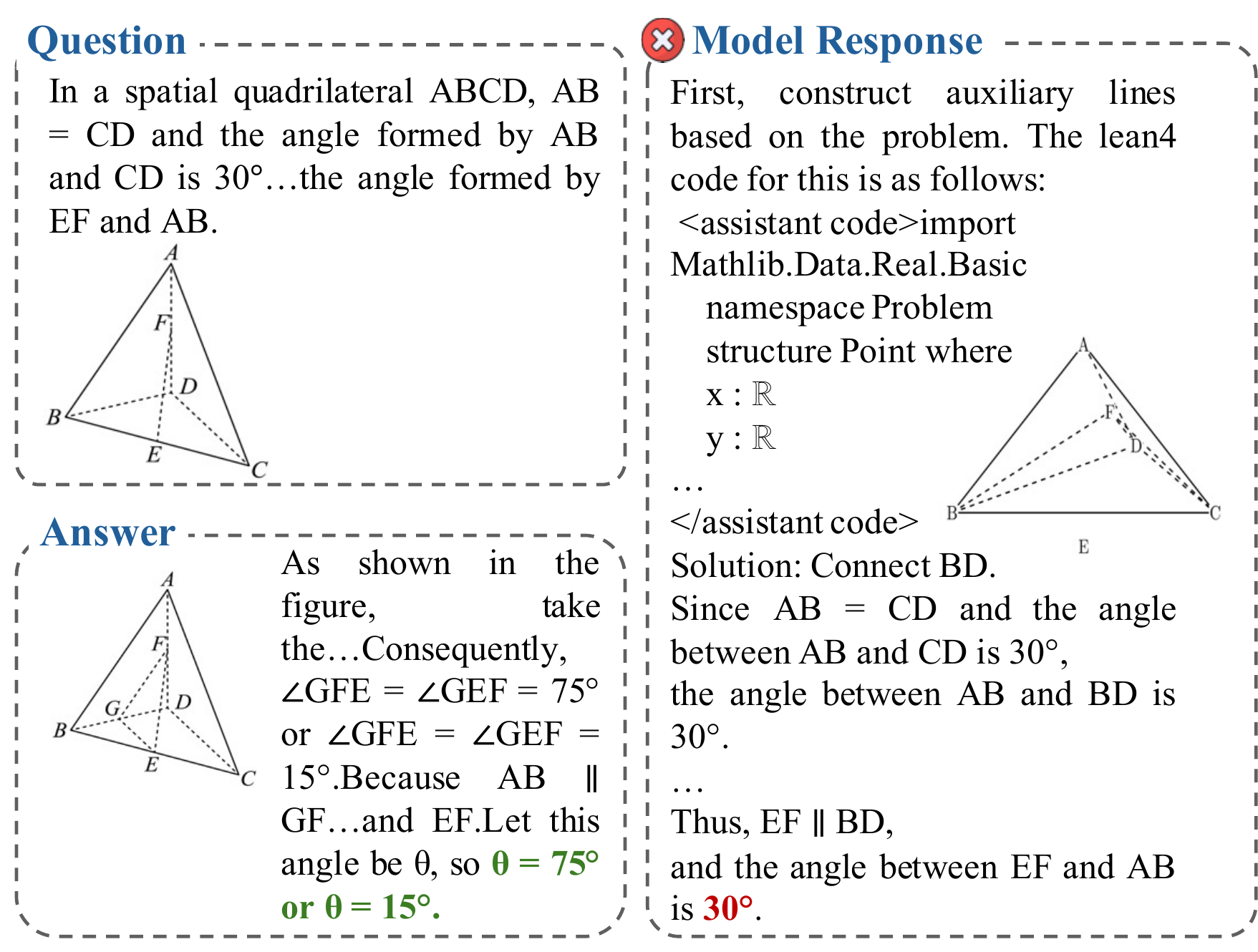}
    \vspace{-5mm}
    \caption{Spatial geometry: mistaken parallelism inference due to midline theorem misapplication.}
    \vspace{-2mm}
    \label{fig:err1}
\end{figure}

While Geoint-R1 achieves strong results overall, several typical errors still arise. As shown in Figure~\ref{fig:err1}, the model may misapply geometric theorems. For example, it wrongly infers that $EF \parallel BD$ simply because $E$ and $F$ are midpoints, neglecting the precise conditions required by the midline theorem and failing to construct key auxiliary points. 

\begin{figure}[h]
    \centering
    \includegraphics[width=\linewidth]{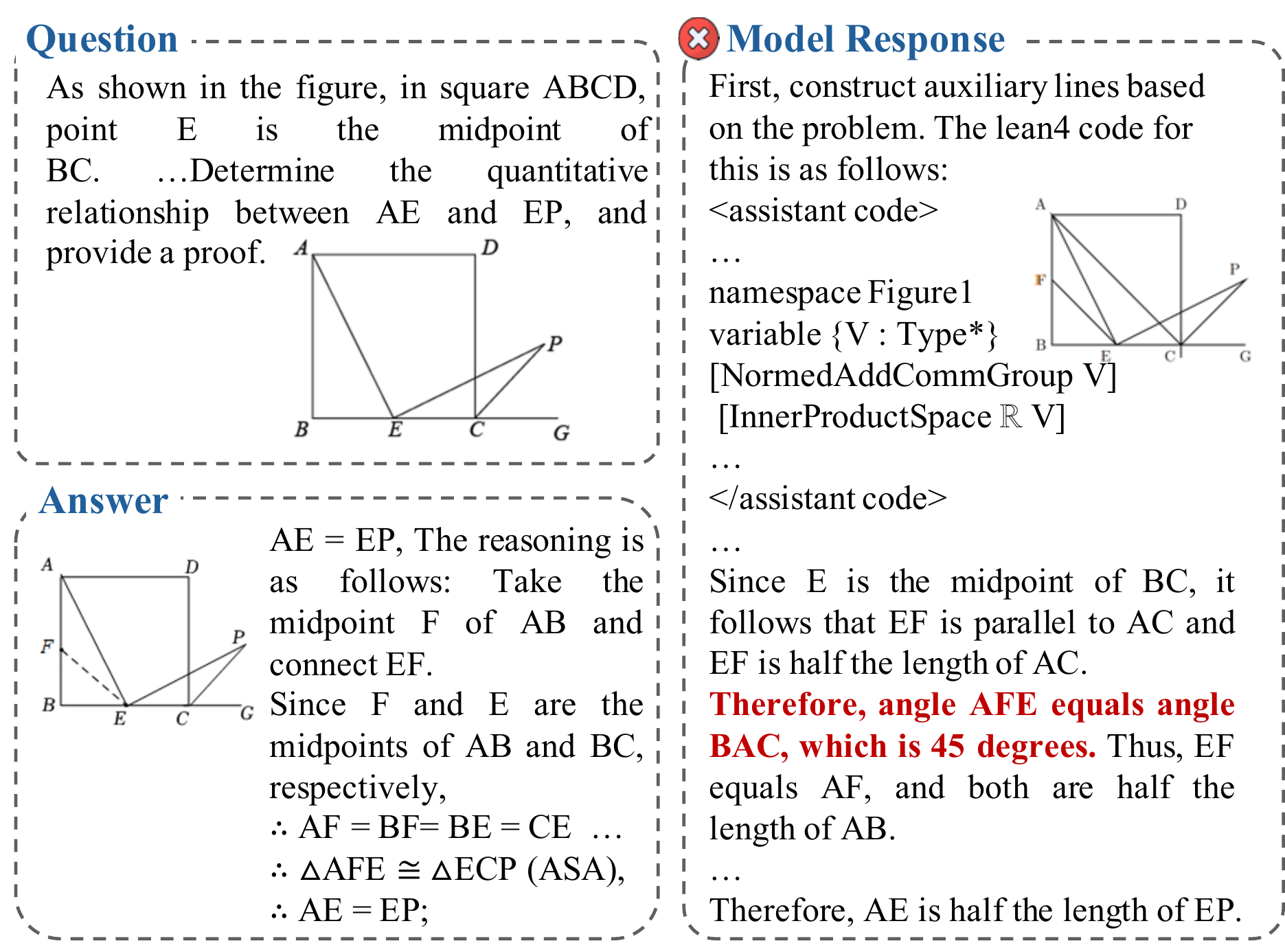}
    \caption{Planar geometry: incomplete reasoning after correct auxiliary construction.}
    \vspace{-2mm}
    \label{fig:err2}
\end{figure}

\begin{figure}[h]
    \centering
    \includegraphics[width=\linewidth]{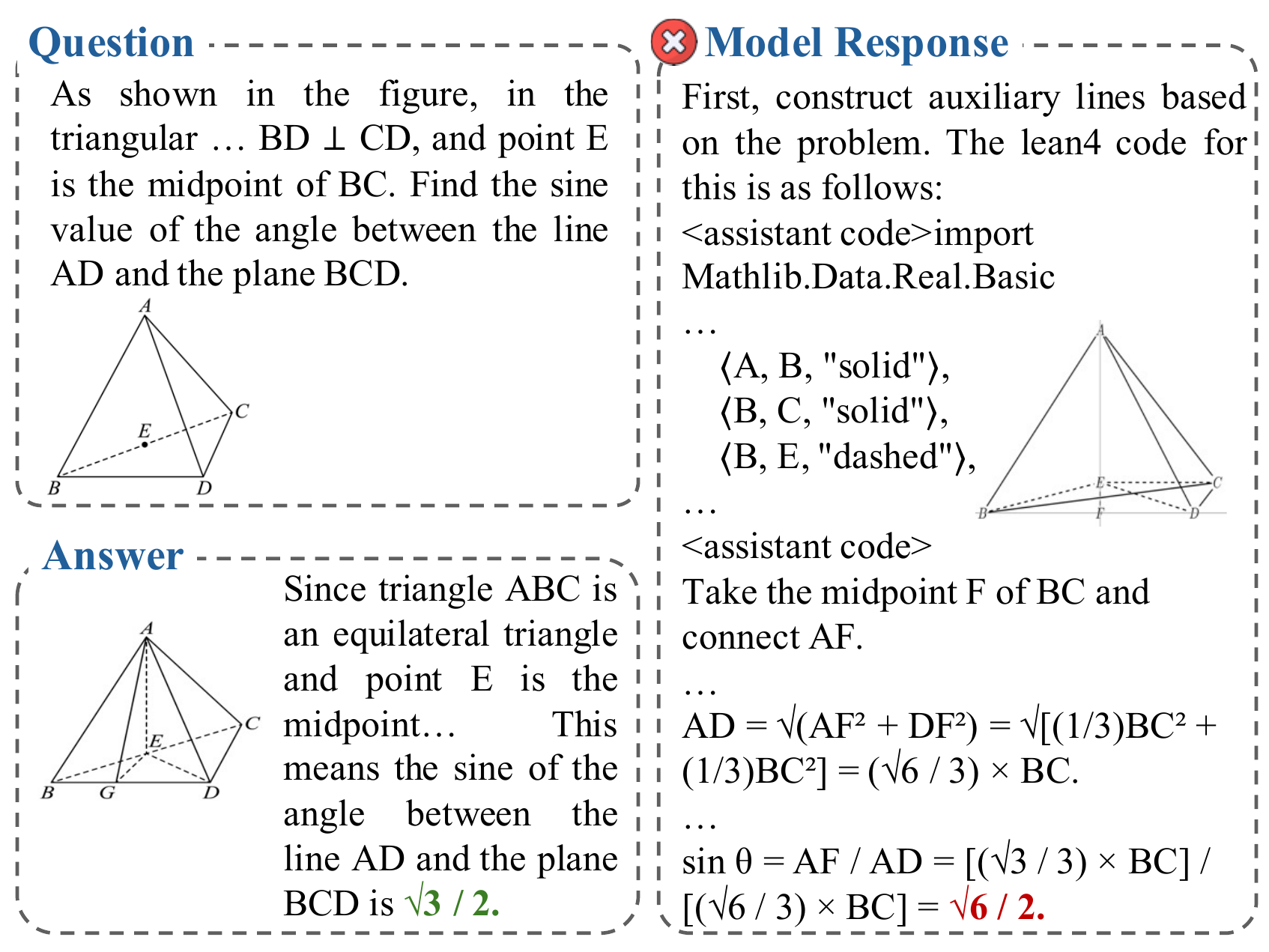}
    \caption{Symbolic confusion and formula error in a multi-step planar problem.}
    \vspace{-4mm}
    \label{fig:err3}
\end{figure}

Figure~\ref{fig:err2} shows a planar geometry case where the model constructs the auxiliary line correctly but fails to establish the right proportional relationships. The logical chain is incomplete, and the reasoning cannot justify the equality between $AE$ and $EP$.
As illustrated in Figure~\ref{fig:err3}, variable confusion and computational mistakes can propagate throughout a multi-step solution. Here, the model mislabels midpoints and ultimately arrives at an impossible value for $\sin \theta > 1$.
These examples demonstrate that Geoint-R1, while powerful, still struggles with strict theorem application, maintaining logical completeness, and tracking symbolic accuracy—especially in complex geometric scenarios.

\section{Conclusion}

In this paper, we introduce Geoint-R1, a multimodal geometric reasoning framework designed to generate formally verifiable solutions with dynamically constructed auxiliary lines. Geoint-R1 integrates supervised fine-tuning, reinforcement learning with verification reward and curriculum learning to effectively handle increasingly complex geometric reasoning tasks. Furthermore, we present the Geoint benchmark, a multimodal geometry dataset incorporating formal Lean4 code annotations, consisting of 1,885 high-quality problems covering diverse geometry categories. Comprehensive experimental evaluations show that Geoint-R1 achieves superior performance over existing state-of-the-art multimodal and mathematical reasoning models, especially excelling in tasks that necessitate explicit auxiliary constructions. Our framework provides a robust foundation for future research in formalized multimodal reasoning and demonstrates significant potential for advancing applications in geometric problem-solving.

\bibliography{aaai2026}


\clearpage
\appendix
\section{Appendix}

\subsection{Evaluation Prompts}
\label{app:evaluate_prompts}

\begin{figure}[h]
    \centering
    \begin{promptbox}[Answer-based Question Grading Prompt]
Now you are playing the role of a strict grading teacher. Your task is to review and score the students' answers based on the standard answers. Throughout the grading process, you need to be familiar with the following key points:

- Grading is only based on the final answers given by the students to determine their correctness, and it does not require checking whether the intermediate solution steps are correct.

- First, extract the final answer from the students' solutions and display it in the analysis results. Then, judge whether the answer is correct.

- Based on your analysis results, give the score. When presenting the scoring basis, you should describe it in segments according to the logic of the analysis. The summary of the scoring basis should be placed at the end, and it can be in the following format: "In conclusion, the student's answer should receive x points" (x represents the specific score of the student).

- Based on your analysis, give the score and display it in a code block in "JSON" format.

Scoring ranges: 1 point if the answer is correct or mathematically equivalent, 0 otherwise.

Output format:

[Grading Basis]:
[Total Score]: float
[JSON]:{
"score": float
}
    \end{promptbox}
    \caption{Evaluation prompt for answer-based questions.}
    \label{fig:qa_prompt}
\end{figure}

\begin{figure}[h]
    \centering
    \begin{promptbox}[Proof-based Question Grading Prompt]
Your task is to evaluate the quality of the proof based on five criteria and give a score between 0 and 1: logical validity (30\%), completeness (20\%), correctness (20\%), construction of auxiliary lines (20\%), and clarity (10\%).

Instructions:
1. Analyze the proof step by step.

2. For each criterion:

- Logical Validity: Check if each step follows logically from the previous one. Flag any logical errors.

- Completeness: Verify if all necessary cases and steps are included to prove the theorem.

- Correctness: Confirm if the final conclusion is correct.

- Construction of auxiliary lines: Determine whether auxiliary lines have been successfully constructed and provide the corresponding Lean4 code for the image.

- Clarity: Assess if the proof is clear, unambiguous, and well-explained.

3. Assign a sub-score (0 to 1) for each criterion and compute the total score using the weights: (0.3 × validity) + (0.2 × completeness) + (0.2 × correctness) + (0.2 × construction) + (0.1 × clarity).

4. Provide a brief explanation (2-3 sentences) summarizing any errors or issues and justifying the score.

Final output format:

[JSON]:
{
"score": float,
"validity": float,
"completeness": float,
"correctness": float,
"construction": float,
"clarity": float,
"explanation": str
}
    \end{promptbox}
    \caption{Evaluation prompt for proof-based questions, following the process-based approach of DeepTheorem~\cite{zheng2024deeptheorem}.}
    \label{fig:proof_prompt}
\end{figure}

\end{document}